\documentclass{article}

\usepackage[final]{neurips_2025}

\usepackage[utf8]{inputenc} 
\usepackage[T1]{fontenc}    
\usepackage{newunicodechar}
\usepackage{hyperref}       
\usepackage{url}            
\usepackage{booktabs}       
\usepackage{amsfonts}       
\usepackage{algorithm}      
\usepackage{nicefrac}       
\usepackage{microtype}      
\usepackage{xcolor}         
\definecolor{myorange}{RGB}{255,165,0}
\definecolor{mygreen}{HTML}{32CD32}
\definecolor{myblue}{RGB}{0,100,255}

\usepackage{amsmath} 
\usepackage{graphicx}
\usepackage{colortbl}
\usepackage{adjustbox}
\usepackage{adjustbox}
\usepackage{ulem}
\usepackage{algpseudocode}

\title{CoFFT: Chain of Foresight-Focus Thought \\for Visual Language Models}

\author{Xinyu Zhang$^{1,2}$
Yuxuan Dong$^{1,2}$
{Lingling Zhang}$^{1,2}$ \thanks{Corresponding author} \;
Chengyou Jia $^{1,2}$ \\ 
\textbf{ZhuoHang Dang} $^{1,2}$
\textbf{Basura Fernando} $^{4,6}$ 
\textbf{Jun Liu}$^{1,3}$ 
\textbf{Mike Zheng Shou} $^{5}$ $^{*}$   \\
{$^{1}$School of Computer Science and Technology, Xi’an Jiaotong University}\; \\
{$^{2}$Ministry of Education Key Laboratory of Intelligent Networks and Network Security, China} \; \\
{$^{3}$Shaanxi Province Key Laboratory of Big Data Knowledge Engineering, China} \; \\
{$^{4}$IHPC, Agency for Science, Technology and Research, Singapore} \; \\
{$^{5}$Show Lab, National University of Singapore} \; \\
{$^{6}$College of Computing and Data Science, Nanyang Technological University, Singapore} \; \\
\texttt{zhang1393869716@stu.xjtu.edu.cn, zhanglling@xjtu.edu.cn, mikeshou@nus.edu.sg}
}

\begin{document}

\maketitle

\begin{abstract}
Despite significant advances in Vision Language Models (VLMs), they remain constrained by the complexity and redundancy of visual input.
When images contain large amounts of irrelevant information, VLMs are susceptible to interference, thus generating excessive task-irrelevant reasoning processes or even hallucinations.
This limitation stems from their inability to discover and process the required regions during reasoning precisely.
To address this limitation, we present the Chain of Foresight-Focus Thought (CoFFT), a novel training-free approach that enhances VLMs' visual reasoning by emulating human visual cognition.
Each Foresight-Focus Thought consists of three stages:
(1) Diverse Sample Generation: generates diverse reasoning samples to explore potential reasoning paths, where each sample contains several reasoning steps;
(2) Dual Foresight Decoding: rigorously evaluates these samples based on both visual focus and reasoning progression, adding the first step of optimal sample to the reasoning process; 
(3) Visual Focus Adjustment: precisely adjust visual focus toward regions most beneficial for future reasoning, before returning to stage (1) to generate subsequent reasoning samples until reaching the final answer.
These stages function iteratively, creating an interdependent cycle where reasoning guides visual focus and visual focus informs subsequent reasoning.
Empirical results across multiple benchmarks using Qwen2.5-VL, InternVL-2.5, and Llava-Next demonstrate consistent performance improvements of 3.1-5.8\% with controllable increasing computational overhead.
\end{abstract}

\section{Introduction}
\label{sec:0}
Vision Language Models (VLMs) have demonstrated remarkable progress across numerous domains \cite{bai2025qwen2, wang2025internvideo2, jia2025chatgen}, particularly in visual reasoning \cite{zhang2025physreason, chen2024spatialvlm, ping2025autogps, huang2025vprochart}.
However, their performance remains significantly constrained by the inherent complexity and redundancy of visual inputs \cite{neo2024towards, yang2024visionzip, zhang2024alignment}.
Visual Language Models demonstrate high sensitivity to large, salient elements in images, often struggling to effectively mitigate the impact of visually dominant but semantically irrelevant information, which leads to deteriorated performance and flawed reasoning \cite{zhang2025mllms}.
These limitations are especially pronounced in complex reasoning tasks that require fine-grained image understanding such as mathematics problem-solving \cite{qiao2024we, he2024olympiadbench}, chart understanding \cite{wang2024charxiv, huang2025evochart}, and geolocation inference \cite{seekworld2025}.
\par
\begin{figure*}[t]
\centering
\includegraphics[width=\textwidth]{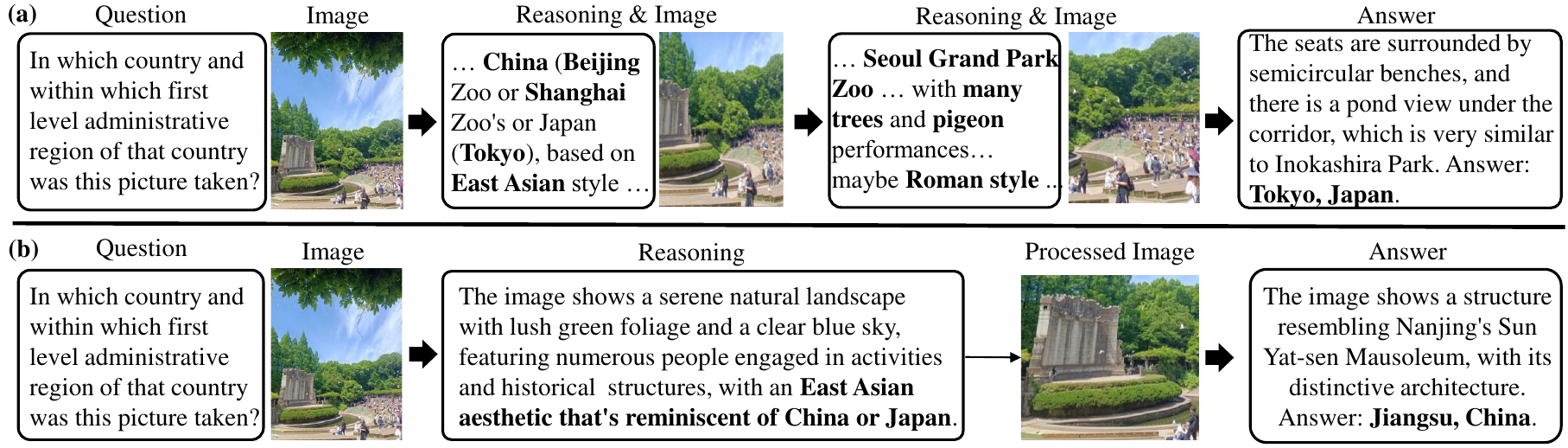}
\vspace{-10pt}
\caption{An example from the SeekWorld \cite{seekworld2025}. (a) is the reasoning process of o3, and (b) is the reasoning process of o3 after human visual cognition. The correct answer is Jiangsu, China.}
\label{fig:0}
\vspace{-20pt}
\end{figure*}

To address this limitation, researchers have proposed Multi-modal Chain of Thought approaches \cite{huvisual, wu2024mind, qi2024cogcom}, which explore the integration of visual operations during reasoning and leverage image information across multiple stages for understanding the image comprehensively.
Taking the recently prominent OpenAI-O3 \cite{openai-o3} as an example, we illustrate its reasoning process as shown in Figure \ref{fig:0} (a).
Despite multiple attempts, O3 ultimately arrives at an incorrect result. 
Based on the redundant and interference-laden reasoning process, we attribute this failure to its tendency to continuously explore different image regions without evaluating their contribution to the question, thus introducing substantial interference from irrelevant visual information.
\par
\par
However, when humans view this image, they evaluate different regions based on their potential contribution to question-solving.
This leads them to reduce attention to the commonplace elements such as crowds, pigeons, and trees while prioritizing attention toward distinctive historical buildings.
We send the region of historical buildings to guide O3, as shown in Figure \ref{fig:0} (b), which illustrates the necessity of learning from human visual cognition.
This performance stems from two human visual cognition capabilities when analyzing complex visual scenes: \textbf{(1) Foresight} - the foresight evaluation of which visual regions will be most valuable for future reasoning \cite{pritchard1960visual}, and \textbf{(2) dynamic visual focus} - precisely shift attention toward the most future reasoning-relevant regions \cite{rayner1995eye,rucci2015control}.
\par
\begin{figure*}[t]
\centering
\includegraphics[width=\textwidth]{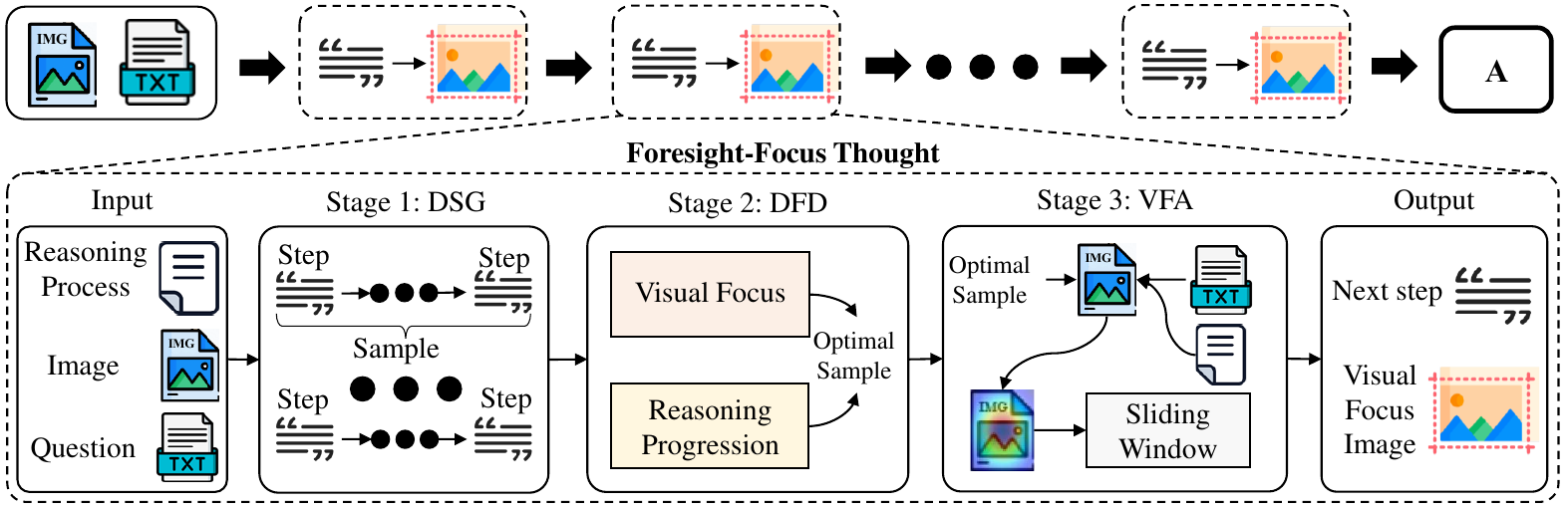}
\vspace{-10pt}
\caption{The overall approach of CoFFT, where Dual Foresight Decoding and Visual Focus Adjustment will be introduced in detail later. }
\label{fig:01}
\vspace{-5pt}
\end{figure*}

Inspired by these observations, we propose the Chain of Foresight-Focus Thought (CoFFT), a training-free approach that enhances VLMs' visual reasoning capabilities, as illustrated in Figure \ref{fig:01}.
Each \textbf{Foresight-Focus Thought} iteration consists of three stages:
\textbf{(1) Diverse Samples Generation (DSG)}: Based on the current reasoning process, the VLM generates diverse reasoning samples under different temperature parameters, where each sample contains multiple subsequent reasoning steps.
\textbf{(2) Dual Foresight Decoding (DFD)}: Evaluates samples by considering both visual focus and reasoning progression to select the optimal reasoning sample, incorporating its first step into the reasoning process.
\textbf{(3) Visual Focus Adjustment (VFA)}: First, the image is evaluated based on two criteria: relevance to the question and correlation with future reasoning steps in the optimal sample.
Then, a sliding window is used to select, crop, and magnify the best region as the next visual focus image.
Finally, this image obtained from VFA serves as visual input for the next iteration, cycling back to stage (1) to generate new reasoning samples with the updated reasoning from stage (2) until reaching the final answer.
This creates a cycle where reasoning guides visual focus, and visual focus informs subsequent reasoning steps, enhancing VLMs' visual reasoning capabilities.
\par
We conduct extensive experiments across several benchmarks using different VLMs, including Qwen2.5-VL \cite{bai2025qwen2}, InternVL-2.5 \cite{wang2025internvideo2}, and Llava-Next \cite{li2024llava}.
CoFFT demonstrates consistent performance gains of 3.1-5.8\% on average across these benchmarks.
Furthermore, analysis across models with varying parameter counts reveals that CoFFT's effectiveness scales positively with model size, yielding greater improvements for larger VLMs.
Our computational overhead analysis demonstrates that while CoFFT requires more computation than direct VLM inference, it remains more efficient than Monte Carlo Tree Search approaches, highlighting the practical efficiency of CoFFT.
\par
\begin{enumerate}
\item We propose CoFFT, a novel training-free approach that improves VLM's performance on complex visual reasoning tasks without requiring model modifications or retraining.
\item We propose Dual-Foresight Decoding, which evaluates reasoning samples by optimizing visual focus and reasoning progression jointly, along with Visual Focus Adjustment that directs visual focus to regions relevant to the question and essential for subsequent reasoning.
\item We demonstrate through extensive experiments across multiple benchmarks that CoFFT significantly improves performance on tasks requiring fine-grained visual understanding and complex reasoning, without excessively increasing computational overhead.
\end{enumerate}
\par

\section{Related work}
\subsection{Visual Language Model}
Large Language Models' (LLMs) success has fundamentally transformed Visual-Language Models (VLMs) development, advancing from basic dispatcher architectures (Visual ChatGPT \cite{wu2023visual}, HuggingGPT \cite{shen2023hugginggpt}, MM-REACT \cite{yang2023mm}) to more sophisticated vision-language approaches.
Key architectural innovations include LLaVA's \cite{liu2023visual} learned image-token projectors, BLIP-2's \cite{li2022blip,li2023blip} question transformers, and MoVA's \cite{zong2024mova} innovative task-specific adaptive routing.
Recent advances feature Qwen2.5-VL's \cite{bai2025qwen2} novel window attention optimizations and InternVL2.5's \cite{wang2025internvideo2} enhanced data processing combining Random JPEG Compression with Square Averaging techniques.

\subsection{Visual Search}
Early research emulated human visual search through Bayesian models with saliency maps\cite{sclar2020modeling, torralba2006contextual}, deep similarity mapping networks\cite{zhang2018finding}, and inverse reinforcement learning\cite{yang2020predicting}. 
These approaches primarily replicated gaze samples but lacked precise target localization capabilities and employed fixed-size attention windows unsuitable for complex scenarios.
Recent methods such as SEAL\cite{wu2024v} incorporate localization modules and visual memory systems to connect visual search with large multimodal models, though requiring additional training. 
DyFo\cite{li2025dyfo} achieves training-free dynamic focus through visual expertise integration, based on language segment-anything modal. 
Both remain limited by their one-time image processing, lacking iterative analytical capabilities during reasoning.
Our approach, by contrast, enables dynamic image manipulation throughout the reasoning process while maintaining a training-free methodology, offering a more robust visual understanding approach.

\subsection{Multi-modal Chain-of-Thought}
Research on multi-modal chain-of-thought reasoning integration follows two main approaches: (1) specialized visual expert models \cite{huvisual, wu2024mind}, such as Sketchpad and VoT/MVoT, which equip VLMs with drawing tools and enhanced spatial reasoning, and (2) improved training processes \cite{yu2025introducing, qi2024cogcom} featuring autonomous region selection and built-in visual operations for preserving reasoning evidence. 
Both approaches face limitations - expert models lack generalizability and require adaptation costs, while enhanced training demands extensive resources. 
Additionally, Multimodal Chain-of-Thought Prompting is an alternative strategy to enhance VLMs' visual reasoning capabilities through prompt engineering \cite{mitra2024compositional,wang2024t,gao2024interleaved}.
However, this method generates only text-based rationales for answers, unlike the aforementioned approaches which can produce interleaved visual-textual rationales.
\section{Method}
\subsection{Overview}
We introduce CoFFT (Chain of Foresight-Focus Thought), a training-free approach designed to address VLM's limitations in complex visual reasoning, as shown in Figure \ref{fig:01}. 
CoFFT is an iterative execution of Foresight-Focus Thought to obtain the final result.
To illustrate the workflow in Foresight-Focus Thought, we take the current $t+1$ iteration of Foresight-Focus Thought as an example, given original image ${V}$, question ${Q}$, current visual focus image $V_{t}$, and existing reasoning process $R_{t}=\{r_1, \dots, r_t\}$, to introduce the following three stages, as shown in Algorithm \ref{alg:cofft}.
\par
\begin{enumerate}
\item \textbf{Diverse Sample Generation}: 
The VLM generates $k$ candidate reasoning samples $S_{t+1}=\{s_1, \dots, s_k\}$ based on the current reasoning process ${R}_t$, visual focus image ${V}_t$, and original question ${Q}$. 
Different temperature parameters are used to get the samples, ensuring sample diversity, where each sample $s$ retains a maximum of $l$ (foresight length) steps.
\item \textbf{Dual Foresight Decoding}:
The samples are evaluated using a combination of visual focus score $E_{att}$ and reasoning progression score $E_{prob}$ to ensure a comprehensive assessment.
$E_{att}$ evaluates the relevance between the reasoning process and image to suppress image-irrelevant hallucination, while $E_{prob}$ assesses reasoning progression by measuring probability improvements across reasoning steps. 
The first step of the optimal sample $s_{i}$ is integrated into the evolving reasoning process ($R_{t+1}$), for continuous iterations.
\item \textbf{Visual Focus Adjustment}: 
First, a scoring mechanism is used to evaluate images based on two criteria: relevance to the question $Q$ and correlation with future reasoning steps in optimal samples $s_{i}$. 
Then, a sliding window selects, crops, and magnifies the highest-scoring region as a visual focus image. Through this stage, the approach switches between global views and local details, avoiding the oversight of critical local information.
\end{enumerate}
\par
\begin{figure*}[t]
\centering
\includegraphics[width=\textwidth]{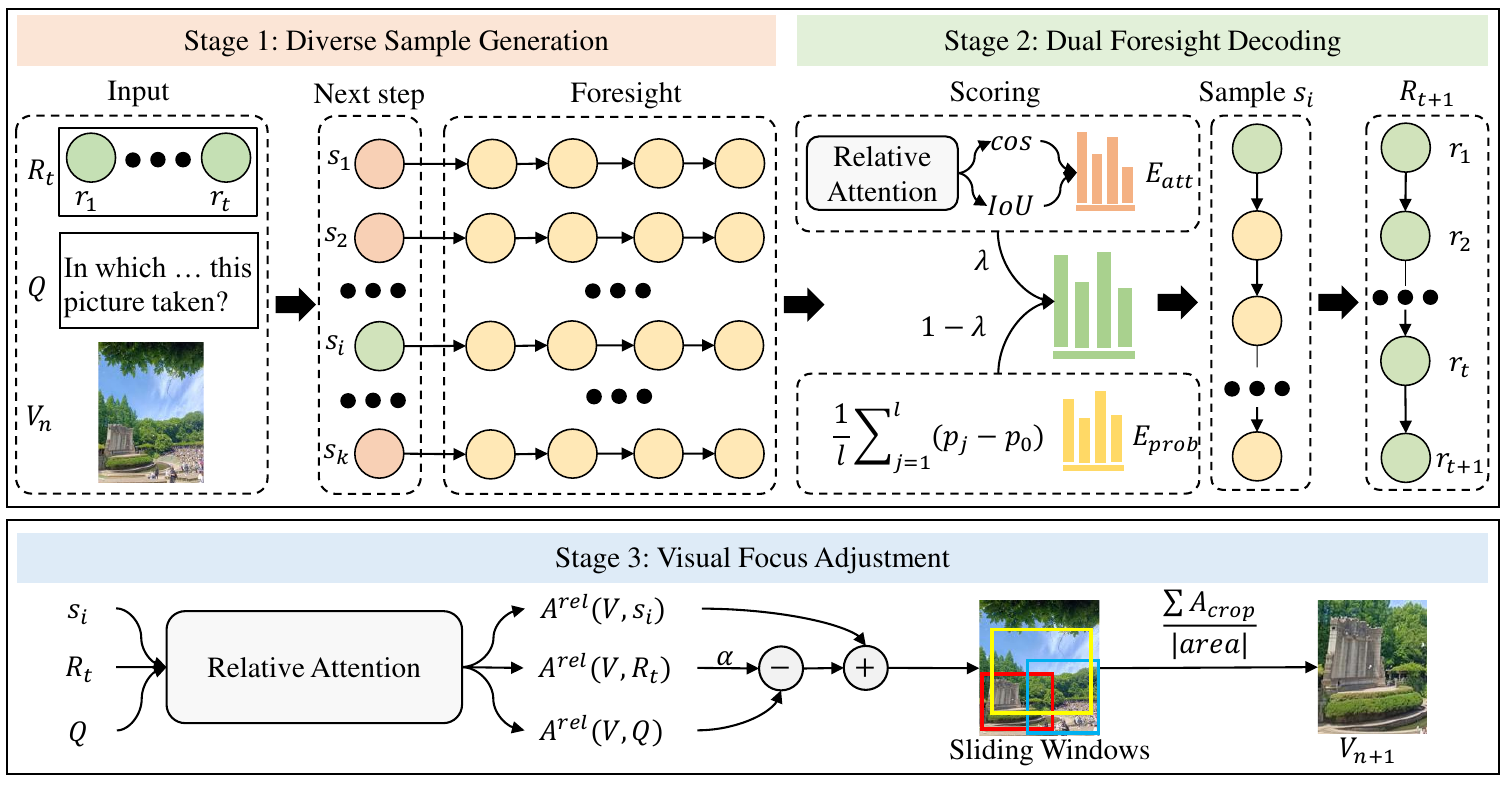}
\vspace{-10pt}
\caption{The two primary components of CoFFT: (a) Dual Foresight Decoding, which evaluates different reasoning samples and selects the best sample from both visual focus and reasoning progression to enhance decision robustness, and (b) Visual Focus Adjustment, which adaptively modulates visual focus adjustment to reasoning-relevant regions for optimized information understanding.}
\vspace{-10pt}
\label{fig:1}
\end{figure*}

These three stages constitute a complete \textbf{Foresight-Focus Thought} iteration as illustrated in Figure \ref{fig:1}.
They create a synergistic cycle where the reasoning path directs visual focus, and optimized visual focus subsequently improves reasoning quality. 
This cognitive-inspired process continues iteratively until the final answer is derived. 
By simulating human visual cognition processes, particularly Dual Foresight Decoding and Visual Focus Adjustment stages, CoFFT reduces VLMs' hallucination tendencies while improving reasoning accuracy. 
Next, we will first introduce the basics of CoFFT, the relative attention mechanism, and then explain the DFD and VFA stages in detail.
\subsection{Relative Attention Mechanism}
Due to the influence of redundancy information in images, relying solely on the original text-image paired attention maps tends to contain noise and makes it challenging to directly distinguish semantically relevant regions \cite{wang2024mllm, zhang2025mllms}.
To address this limitation, we introduce a relative attention mechanism that normalizes any text input attention against a baseline descriptive attention distribution.
\par
Formally, taking question $Q$ and image $V$ as an example, we define relative attention $A^{rel}({V}, {Q})$ as:
\begin{equation}
A^{rel}({V}, {Q}) = Softmax(\frac{A({V}, {Q})}{A({V}, {D}) + \epsilon}), \quad A \in \mathbb{R}^{H \times W}
\label{eq:0}
\end{equation}
where $A({V}, {Q})$ and $A({V}, {D})$ represent the attention maps for the question and descriptive prompts (e.g., ``Describe the image in detail'') respectively, the $Softmax$ function is used to normalize, with $\epsilon$ ($10^{-10}$) ensuring numerical stability. 
The division operation is performed element-wise.
 It is worth noting that the attention map $A$ between the input text and the image is provided by VLM itself and does not require additional modification during reasoning process.
\par
This relative attention mechanism is effective for two reasons: (1) it suppresses attention to generally salient but input text-irrelevant regions by normalizing against the descriptive attention, and (2) it amplifies attention to regions that are specifically relevant to the input text, thereby encouraging the model to focus on regions that are relevant to the input text rather than salient elements in the image.
\par
\subsection{Dual-Foresight Decoding}
Our Dual-Foresight Decoding mechanism evaluates reasoning samples by jointly combining visual focus and reasoning progression. 
Given a set of candidate samples $S_{t+1}$ generated at reasoning step $t+1$, we evaluate each sample $s \in S_{t+1}$ (containing up to $l$ steps) through a combined score:
\begin{equation}
E_{t+1} = \lambda \cdot Softmax([E_{att}(s)]\;{\forall s \in S_{t+1}}) + (1-\lambda) \cdot Softmax([E_{prob}(s)]\;{\forall s \in S_{t+1}}),
\end{equation}
where $\lambda \in (0,1]$ (empirically set to 0.3) balances visual focus and reasoning progression. 
The $Softmax$ function normalizes scores across all samples to obtain comparable distributions.
\par
For the visual focus score, we design $E_{att}$ that measures the focus between each sample and the image, which comprehensively integrates cosine similarity and IoU metrics for robust evaluation::
\begin{equation}
E_{att}(s) = 0.5 \cdot \cos(A^{rel}({V}, {Q}), A^{rel}(V, s)) + 0.5 \cdot {IoU}^{30\%}(A^{rel}({V}, {Q}), A^{rel}(V, s)),
\end{equation}
where $A^{rel}(V, s)$ computes the relative attention following Equation \ref{eq:0}, and ${IoU}^{30\%}$ calculates the intersection over union of high-attention regions (top 30\%). 
This dual-metric formulation captures both global attention distribution and local focus similarity, effectively reducing hallucination by enforcing visual coherence through comprehensive spatial analysis.
\par
Inspired by \cite{ma2024non, xu2025phi}, we observe that scores based on dynamic improvement tend to be more reliable than those simply using confidence values.
Therefore, we propose the reasoning progression score $E_{prob}$ that captures the stepwise improvement in reasoning quality using new reasoning steps of different lengths to compare with the current reasoning process:
\begin{equation}
E_{prob}(s) = \frac{1}{l} \sum_{j=1}^{l} (p_j - p_{0}),
\end{equation}
where $p_j$ represents the mean log probability of tokens including $R_{t}$ and first $j$-th step of sample $s$, and $l$ denotes the number of evaluated steps.
This formulation identifies samples demonstrating maximum average improvement in reasoning confidence, leading to more reliable sample selection.
\subsection{Visual Focus Adjustment}
This stage dynamically adjusts visual focus by evaluating and selecting informative image regions based on a dual-criteria scoring mechanism: question relevance and future reasoning relevance. 
This enables effective switching between global views and local details throughout the reasoning process.
\par
For question relevance, we encourage VLM to focus on regions that have not been explored by the current reasoning process $R_t$. Therefore, we define $C^{rel}({V}, {Q}, {R}_t)$ as following:
\begin{equation}
C^{rel}({V}, {Q}, {R}_t) = {max}(A^{rel}({V}, {Q}) - \alpha \cdot A^{rel}({V}, {R}_t), 0).
\end{equation}
where $A^{rel}({V}, {R}_t)$ represents the relative attention from $n$ completed reasoning steps (following Equation \ref{eq:0}), and $\alpha \in (0,1]$ (empirically set to 0.3) controls the influence of processed regions.
\par
For future reasoning relevance, we leverage the optimal sample $s_i$ selected by the Dual Foresight Decoding stage to obtain $A^{rel}({V}, s_i)$ in the same way as Equation \ref{eq:0}.
\par
Based on those, the final attention map evaluation combines both criteria as following:
\begin{equation}
A_{crop}= 0.5 \cdot C^{rel}({V}, {Q}, {R}_t) + 0.5 \cdot A^{rel}({V}, s_i).
\end{equation}
\par
Then, a dynamic sliding window algorithm determines the optimal visual region ${V}_{t+1}$:
\begin{equation}
{V}_{t+1} =
\begin{cases}
\arg\max_{B \in \Omega} \frac{1}{|B|} \sum_{(x,y) \in B} A_{crop}(x,y), & \text{if } \mu_{B^{*}} > \mu_{{V}_0} + \beta \\
{V}_0, & \text{otherwise}
\end{cases}
\end{equation}
where $\beta = \sigma_{V_0} \cdot (1 - \cos(C^{rel}({V}, {Q}, {R}_t), A^{rel}({V}, s_i)))$, and $B^{*}$ is the optimal region. 
Here, $\sigma_{V_0}$ denotes the standard deviation of $A_{crop}$, $\Omega$ contains regions spanning 40\%-90\% (interval is 10\%) of original dimensions, and $\mu_{B^{*}}=\frac{1}{|B^{*}|} \sum_{(x,y) \in B^{*}} A_{crop}(x,y)$ and $\mu_{{V}_0}=\frac{1}{|{V}_0|} \sum_{(x,y) \in {V}_0} A_{crop}(x,y)$ represent the maximum region and global mean scores. 
The selected region is scaled maintaining an aspect ratio within the original dimensions for subsequent reasoning iterations.
\par
The mechanism employs adaptive thresholding based on the two attention maps: requiring minimal improvement over the global mean when attention centers converge ($\beta \approx 0$) and stricter thresholds when they diverge ($\beta \approx \sigma_{V_0}$). 
This strategy prevents fixation on misleading local information while ensuring comprehensive visual understanding through balanced global-local focus transitions.
\section{Experiment}
\label{sec:4}
\subsection{Settings}
\label{sec:4.1}
\paragraph{Benchmarks}
We conduct comprehensive evaluations across multiple complementary benchmarks to assess various aspects of visual reasoning capabilities. 
For mathematical and geometric reasoning, we employ MathVista~\cite{lumathvista} and MathVision~\cite{wang2024measuring}. 
To evaluate cross-domain visual reasoning abilities, we utilize the multi-subject benchmarks M3CoT~\cite{chen2024m3cot} and MMStar~\cite{chenwe}. For chart comprehension assessment, we leverage Charxiv~\cite{wang2024charxiv}. 
Additionally, we contribute to the geographical domain by introducing two novel datasets in SeekWorld~\cite{seekworld2025}: SeekWorld-Global, which utilizes Google Maps panoramic imagery, and SeekWorld-China, which incorporates data from the Xiaohongshu App. 
\begin{table}[t]
\centering
\caption{Main results, where Claude-3.5, Gemini-2, Pred. Dec. are Claude-3.5-sonnet, Gemini-2.0-Flash and Predictive decoding. The optimal results are highlighted in bold, whereas suboptimal results are underlined. The Avg. column indicates the averaged results across the six benchmarks.}
\begin{adjustbox}{width=\textwidth, max width=\textwidth}
\begin{tabular}{l|cc|cc|c|cc|c}
\toprule
& \multicolumn{2}{c|}{\textbf{Math}} & \multicolumn{2}{c|}{\textbf{Multi-subjects}} & \textbf{Chart} & \multicolumn{2}{c|}{\textbf{Geography}} & \\
\textbf{Models} & {MathVista} & {MathVision} & {MMStar} & {M3CoT} & {Charxiv} & {S.W-China} & {S.W-Global} & {Avg.} \\
\midrule
\rowcolor{gray!20} \multicolumn{9}{c}{\textbf{Closed source VLMs}} \\
GPT-4o & 63.8 & 18.75 & 64.7 & 65.75 & 50.5 & 31.90 & 56.50 & 50.27 \\
Claude-3.5 & 65.4 & 26.21 & 65.1 & 66.05 & 60.2 & 29.22 & 52.50 & 52.10\\
Gemini-2 & 73.1 & 31.83 & 69.4 & 67.73 & 53.2 & 30.83 & 55.31 & 54.49\\
\midrule
\rowcolor{gray!20} \multicolumn{9}{c}{\textbf{Qwen2.5VL-7B-Instruct}} \\
Baseline & 68.2 & 18.09 & 63.9 & 59.62 & 42.5 & 21.45 & 25.31 & 42.72\\
MCTS & 69.6 & 18.75 & \underline{66.2} & 60.87 & 44.5 & 26.27 & 26.56 & 44.68\\
Pred. Dec. & \underline{69.9} & \underline{19.73} & 65.7 & \underline{61.34} & \underline{45.3} & 26.54 & 26.86 & \underline{45.05}\\
ICoT & 47.5 & 10.53 & 42.1 & 61.17 & 27.5 & 29.22 & 26.37 & 34.91\\
DyFo & 68.4 & 16.78 & 64.5 & 61.26 & 43.7 & \underline{32.44} & \underline{27.81} & 44.98 \\
CoFFT & \textbf{70.4} & \textbf{23.36} & \textbf{69.4} & \textbf{62.47} & \textbf{47.2} & \textbf{35.12} & \textbf{29.37} & \textbf{48.19}\\
\midrule
\rowcolor{gray!20} \multicolumn{9}{c}{\textbf{LLaVA-NeXT-7B}} \\
Baseline & 34.6 & 9.87 & 34.2 & 38.27 & 13.9 & 10.72 & 15.31 & 22.41\\
MCTS & \underline{35.1} & 10.53 & \underline{36.7} & 39.52 & 14.6 & 12.33 & 16.56 & 23.62\\
Pred. Dec. & 34.8 & \underline{11.36} & 35.1 & \underline{40.16} & 15.3 & 11.80 & 17.19 & 23.67\\
ICoT & 27.3 & 11.18 & 31.2 & 39.34 & 9.5 & 12.87 & 16.56 & 21.14 \\ 
DyFo & 34.8 & 8.22 & 36.1 & 39.86 & \underline{15.7} & \underline{13.67} & \underline{17.50} & \underline{23.69}\\
CoFFT & \textbf{35.6} & \textbf{12.17} & \textbf{38.3} & \textbf{40.68} & \textbf{16.8} & \textbf{15.55} & \textbf{19.69} & \textbf{25.54}\\
\midrule
\rowcolor{gray!20} \multicolumn{9}{c}{\textbf{InternVL2.5-8B-Instruct}} \\
Baseline & 64.4 & 22.00 & 60.5 & 57.16 & 32.9 & 23.32 & 25.63 & 40.84\\
MCTS & 65.0 & 24.67 & 62.0 & 58.24 & 34.3 & 25.20 & 26.56 & 42.28 \\
Pred. Dec. & \underline{65.4} & \underline{25.00} & \underline{62.4} & \underline{58.76} & \underline{35.1} & 26.81 & 27.19 & \underline{42.95}\\
ICoT & 42.9 & 16.12 & 39.4 & 58.50 & 16.4 & 27.35 & 26.88 & 32.51\\
DyFo & 64.7 & 20.39 & 61.5 & 58.58 & 34.2 & \underline{29.22} & \underline{28.13} & 42.39 \\
CoFFT & \textbf{66.5} & \textbf{28.29} & \textbf{64.5} & \textbf{59.19} & \textbf{36.6} & \textbf{31.37} & \textbf{30.63} & \textbf{45.30} \\
\midrule
\rowcolor{gray!20} \multicolumn{9}{c}{\textbf{Qwen2.5VL-32B-Instruct}} \\
Baseline & 74.7 & 25.33 & 69.5 & 62.81 & 44.5 & 24.13 & 28.41 & 47.05\\
MCTS & 76.2 & 27.31 & 70.6 & 64.15 & 47.6 & 28.69 & 29.06 & 49.09\\
Pred. Dec. & \underline{76.6} & \underline{27.96} & \underline{71.1} & \underline{64.62} & \underline{48.2} & 29.22 & 30.31 & 49.72 \\
ICoT & 58.7 & 21.05 & 54.6 & 63.93 & 47.3 & 32.17 & 31.56 & 44.19 \\
DyFo & 75.6 & 24.67 & 70.1 & 64.32 & 47.7 & \underline{35.38} & \underline{32.19} & \underline{49.99} \\
CoFFT & \textbf{77.5} & \textbf{29.93} & \textbf{72.7} & \textbf{66.08} & \textbf{50.9} & \textbf{38.61} & \textbf{34.38} & \textbf{52.96}\\
\bottomrule
\end{tabular}
\end{adjustbox}
\label{tab:0}
\end{table}

\begin{table}[t]
\centering
\caption{FLOPS denotes the calculated computational cost, with lower values indicating lower costs.}
\begin{tabular}{l|cccccc}
\toprule
\textbf{Models} & {Baseline}& {MCTS} & {Predictive decoding}  & {ICoT} & {DyFo} & {CoFFT} \\
 \midrule
FLOPS & 8.35e+12 & 4.05e+14 & 1.85e+14 & 1.88e+13 & 1.98e+13 & 2.38e+14\\
\bottomrule
\end{tabular}
\label{tab:10}
\vspace{-10pt}
\end{table}
\paragraph{Comparison methods and experimental setup}
Our experimental approach incorporates state-of-the-art Vision Language Models (VLMs): Qwen2.5-VL-Instruct (7B, 32B) \cite{bai2025qwen2}, InternVL2.5-Instruct (8B) \cite{wang2025internvideo2}, and Llava-Next (7B) \cite{li2024llava}, selected for their architectural capabilities and superior performance in visual reasoning.
We establish comprehensive baseline comparisons using search-based method (MCTS \cite{coulom2006efficient}), foresight reasoning method (Predictive Decoding \cite{ma2024non}), visual search methodology (DyFo \cite{li2025dyfo}), and multi-modal chain-of-thought prompting method (ICoT \cite{gao2024interleaved}).
This diverse selection enables systematic evaluation against both text-based reasoning, visual search and multi-modal chain-of-thought methods.
All experiments are run on four NVIDIA A100 GPUs with parallel processing. 
To ensure sample diversity, the temperature parameter ranges from $0.4$ to $1$ with an interval of $0.1$, and a sample is randomly selected each time it is generated. 
To prevent repeated selections, the probability weight of each chosen parameter is reduced by half in subsequent sampling processes. 
The weights are reset to their initial values once all parameters have been selected, ensuring a balanced exploration of different temperature values.
For Predictive Decoding and CoFFT, inference is considered complete when the model outputs `REASONING\_COMPLETE'.
\paragraph{Performance Metrics}
We adopt Pass@1 accuracy (Acc.) as our primary performance metric across all benchmarks. 
To quantify the computational efficiency-performance trade-off, we calculate floating point operations (FLOPS) following the methodology in \cite{kaplan2020scaling}, where $\text{FLOPS} \approx 6nP$ ($P$ represents model parameters, $n$ denotes generated tokens). 
By computing the average number of tokens generated per example, we provide a standardized measure of computational cost across different methods based on Qwen2.5-VL-7B-Instruct to enable direct efficiency comparisons.

\subsection{Results}
\label{sec:4.2}
As shown in Tables \ref{tab:0} and \ref{tab:10}, we compare our method against current state-of-the-art approaches across seven datasets and several VLMs. 
Our analysis reveals several significant findings:
\paragraph{Comprehensive Performance Improvements}
CoFFT consistently outperforms all approaches across all VLMs and seven benchmarks, as demonstrated in Table \ref{tab:0}.
Performance gains are substantial across diverse model scales, from 7B to 32B parameters, highlighting CoFFT's versatility and effectiveness in enhancing both visual perception and reasoning capabilities of foundation models.
\paragraph{Balancing Visual Understanding and Reasoning}
VLMs must balance image comprehension with reasoning capabilities. Language-based methods like MCTS and Predictive Decoding show broad improvements but struggle with fine-grained visual analysis tasks in the SeekWorld datasets, indicating single-pass image understanding is often insufficient.
Similarly, DyFo and ICoT fail to significantly improve performance on reasoning-intensive tasks, primarily due to the fragility of the reasoning process.
Our proposed CoFFT addresses these limitations by simultaneously enhancing both visual comprehension and reasoning capabilities, achieving superior performance.
\paragraph{Fine-grained Detail Extraction}
CoFFT excels particularly in extracting easily overlooked fine-grained details from images, evidenced by substantial improvements on Charxiv, SeekWorld-China, and SeekWorld-Global benchmarks. 
Nevertheless, on SeekWorld-Global, our method still underperforms compared to advanced closed-source models like GPT-4o and Gemini-2.0-Flash.
We attribute this gap to differences in knowledge capabilities across regional contexts, where these closed-source models likely benefit from more extensive pre-training on globally diverse datasets.
\paragraph{Scaling Properties}
Notably, the benefits of CoFFT increase with model size in a systematic manner.
The absolute performance improvements are more pronounced with larger models, suggesting that CoFFT effectively leverages the enhanced capabilities of more powerful base models. 
For instance, comparing improvements on Qwen2.5VL-7B-Instruct versus Qwen2.5VL-32B-Instruct reveals larger absolute gains across most benchmarks for the 32B model. 
This scaling trend suggests CoFFT could yield even greater benefits when applied to larger foundation models.
\paragraph{Computational Efficiency Analysis}
As shown in Table \ref{tab:10}, while CoFFT introduces additional computational costs compared to the baseline, it remains more computationally efficient than MCTS while delivering superior performance across all benchmarks. 
The computational overhead primarily stems from the iterative reasoning process. 
Given the significant performance improvements observed, particularly on challenging reasoning tasks, this computational trade-off is well justified.

\begin{table}[t]
\centering
\caption{Ablation Studies on Qwen2.5VL-7B-Instruct. \textit{w/o DFD} refers to sample evaluation using reasoning progression score only.
\textit{w/o VFA} indicates the absence of adaptive image cropping, relying instead on original images throughout the reasoning process.}
\begin{adjustbox}{width=\textwidth, max width=\textwidth}
\begin{tabular}{l|cc|cc|c|cc}
\toprule
& \multicolumn{2}{c|}{\textbf{Math}} & \multicolumn{2}{c|}{\textbf{Multi-subjects}} & \textbf{Chart} & \multicolumn{2}{c}{\textbf{Geography}} \\
\textbf{Models} & {MathVista} & {MathVision} & {MMStar} & {M3CoT} & {Charxiv} & {S.W-China} & {S.W-Global} \\
\midrule
Our & \textbf{70.4} & \textbf{23.36} & \textbf{69.4} & \textbf{62.47} & \textbf{47.2} & \textbf{35.12} & \textbf{29.37} \\
w/o DFD & 68.5 & 20.42 & 66.5 & 61.39 & 44.8 & 28.42 & 27.19 \\
w/o VFA & 69.3 & 21.71 & 67.4 & 61.09 & 44.7 & 27.08 & 26.25 \\
\bottomrule
\end{tabular}
\end{adjustbox}
\label{tab:1}
\end{table}

\begin{table}[t]
\centering
\caption{Performance comparison of method combinations on Qwen2.5-VL-7B. We evaluate various combinations of existing approaches (DyFo, Predictive Decoding) and our proposed components (VFA, DFD) to demonstrate the effectiveness of our integrated CoFFT framework.}
\begin{tabular}{lcccc}
\hline
\textbf{Method/Combination} & \textbf{MathVista} & \textbf{SeekWorld-China} & \textbf{Average}  \\
\hline
Qwen2.5-VL-7B (Baseline) & 68.2 & 21.45 & 44.83 \\
DyFo + Predictive Decoding & 69.0 & 33.24 & 51.02 \\
VFA + Predictive Decoding & 68.5 & 31.37 & 50.24 \\
DyFo + Dual Foresight Decoding & 69.3 & 34.05 & 51.73  \\
\textbf{CoFFT (VFA + DFD)} & \textbf{70.4} & \textbf{35.12} & \textbf{52.76} \\
\hline
\end{tabular}
\label{tab:1111}
\end{table}
\subsection{Ablation Studies}
To validate the effectiveness of our proposed components, we conduct ablation experiments across all benchmarks using Qwen2.5-VL-7B-Instruct as the backbone model. 
We specifically examine two critical components: \textit{w/o DFD} removes the visual focus score $E_{att}$ and relies solely on the reasoning progression score $E_{prob}$, while \textit{w/o VFA} eliminates the dynamic image update module and uses only the original image for inference during the reasoning process. 
Results in Table \ref{tab:1} demonstrate performance degradation across all benchmarks when either component is removed, confirming the integral contribution of each mechanism to the overall effectiveness of CoFFT.
\par
Additionally, we explore combinations of existing methods such as DyFo for visual search and Predictive Decoding for reasoning enhancement, with results shown in Table \ref{tab:1111}. 
Our evaluation demonstrates that CoFFT achieves superior performance with an overall accuracy of 48.19\% compared to the baseline accuracy of 42.72\%. Experiments reveal that simply combining existing visual search and reasoning methods often yields suboptimal results due to fundamental incompatibilities between their design principles. 
DyFo's static region determination conflicts with Predictive Decoding's reasoning requirements, while VFA's dynamic adjustments cannot be effectively utilized by static reasoning approaches. 
CoFFT's breakthrough lies in its synergistic design where DFD evaluates reasoning paths based on both logical progress and visual relevance to guide VFA's region selection, while VFA provides high-quality visual inputs that enable DFD to make more reliable judgments. 
This mutual reinforcement between focus and reasoning represents CoFFT's core contribution and explains its significant performance improvements over alternative combinations.

\begin{table}[t]
\centering
\caption{Analysis of the effects of Foresight parameter $l$ and Sampling number K on experimental outcomes. 
During $k$ parameter studies, $l$ is maintained constant at $5$, while $l$ parameter studies are conducted with $k$ fixed at $4$.}
\begin{adjustbox}{width=\textwidth, max width=\textwidth}
\begin{tabular}{c|ccc|c|ccc}
\toprule
\textbf{Foresight } & \textbf{MathVista} & \textbf{S.E-China} & \textbf{FLOPS} & \textbf{Sampling } & \textbf{MathVista} & \textbf{S.E-China} & \textbf{FLOPS}\\
\midrule
$l=3$ & 68.8 & 33.51 & 1.41e+14 & $k=2$ & 68.8 & 34.32 & 1.19e+14\\
$l=4$ & 69.3 & 34.58 & 1.91e+14 & $k=4$ & 70.4 & 35.12 & 2.38e+14\\
$l=5$ & 70.4 & 35.12 & 2.38e+14 & $k=6$ & 70.8 & 35.65 & 3.56e+14\\
$l=6$ & 70.5 & 35.65 & 2.86e+14 & $k=8$ & 71.4 & 36.19 & 4.75e+14\\
$l=7$ & 70.7 & 35.92 & 3.33e+14 & $k=10$ & 72.2& 37.27 & 5.93e+14\\
\bottomrule
\end{tabular}
\label{tab:2}
\end{adjustbox}
\label{tab:model_performance}
\vspace{-10pt}
\end{table}

\begin{figure*}[t]
\centering
\includegraphics[width=\textwidth]{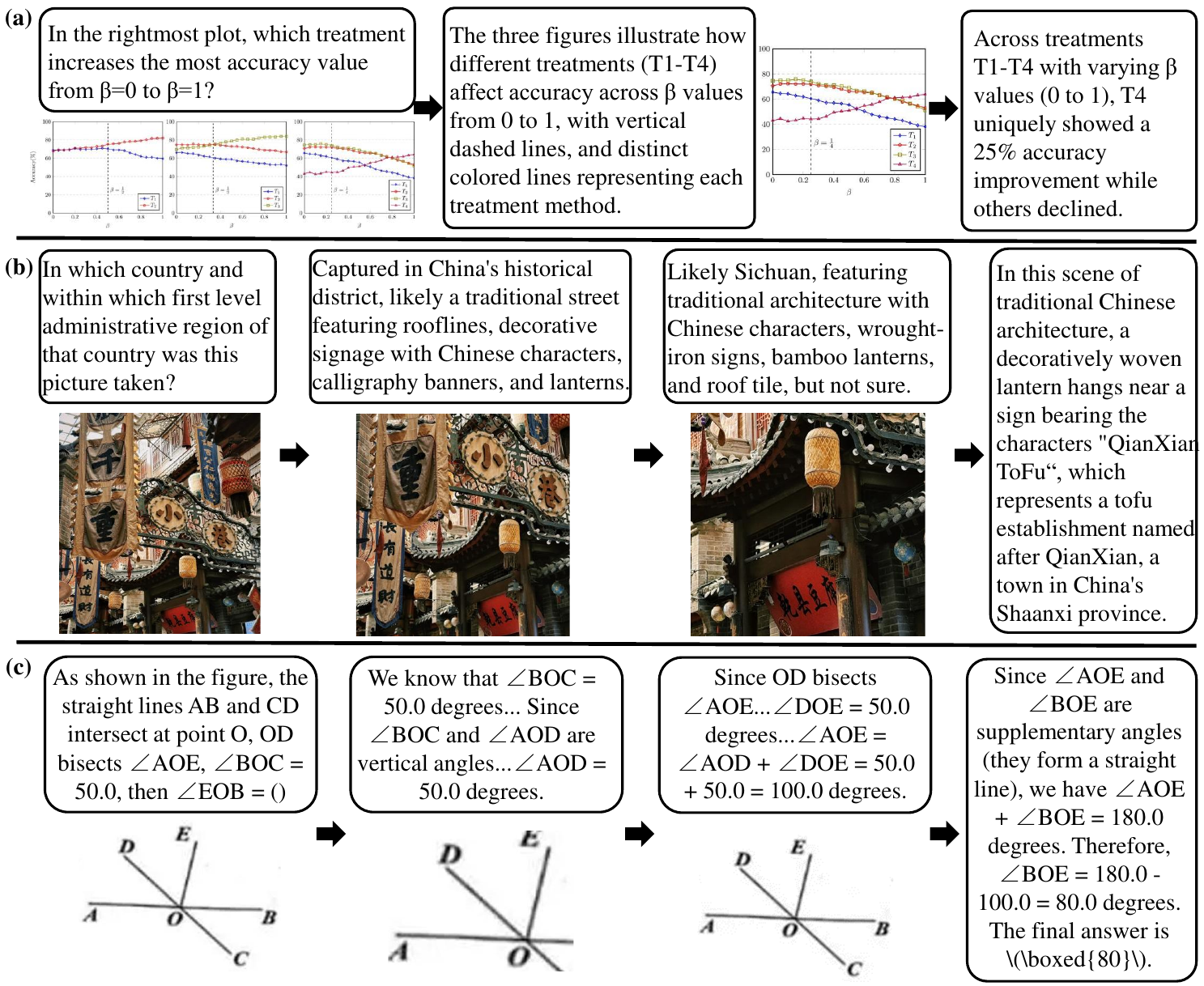}
\vspace{-10pt}
\caption{Illustrative cases demonstrating the reasoning process of CoFFT. (a), (b), and (c) show an example from Charxiv, SeekWorld-China, and MathVista benchmarks, respectively. }
\label{fig:2}
\vspace{-10pt}
\end{figure*}
\subsection{Parameter Analysis}
Our CoFFT approach incorporates two critical hyperparameters: the number of foresight statements ($l$) and the number of samples generated per inference step ($k$). 
For our primary experiments, we select $l=5$ and $k=4$ to balance computational efficiency and performance.
To investigate parametric sensitivity, we systematically varied $l$ from $3$ to $7$ (interval is $1$) and $k$ from $2$ to $10$ (interval is $2$) using Qwen2.5-VL-Instruct as our backbone model.
Experiments on MathVista and SeekWorld-China, which represent different visual reasoning aspects, show consistent performance gains with increasing $l$ and $k$ (Table \ref{tab:2}). This scalability suggests potential for further improvements with larger parameters, though practical deployment requires balancing performance with computational costs.

\subsection{Case Study}
Figure \ref{fig:2} illustrates the performance of CoFFT through three different cases.
In case (a), CoFFT successfully identifies relevant chart to correctly answer this question.
In case (b), despite the presence of substantially irrelevant information in the complete image, CoFFT exhibits remarkable discrimination by precisely identifying and focusing on critical regions, thereby extracting essential information that leads to accurate results.
In case (c), CoFFT successfully identifies relevant regions while demonstrating the ability to dynamically return to the original image when necessary for continued reasoning.
These cases demonstrate the approach's adaptability across varying visual complexity levels and information densities, highlighting its robustness in real-world applications.

\section{Conclusion}
\label{sec:5}
We have introduced CoFFT, a training-free approach that enhances visual reasoning capabilities in VLMs by emulating human visual cognition through three stages: Diverse Sample Generation, Dual-Foresight Decoding, and Visual Focus Adjustment.
CoFFT effectively bridges the gap between static visual processing and dynamic reasoning by implementing a sophisticated visual cognition mechanism that continuously refines visual focus based on comprehensive iterative reasoning outputs.
Extensive experiments demonstrate a significant 3.1\%-5.8\% average improvement across challenging visual reasoning tasks without requiring specialized visual expert models or system modifications.
However, while CoFFT successfully solves previously unsolvable problems for VLMs, it may introduce unexpected errors in cases where VLMs alone perform well, indicating potential interference with their original well-established capabilities.
This suggests that although CoFFT represents a promising approach for VLM enhancement, its robustness requires further improvement, making the systematic optimization of the reasoning process an important direction for future research.

\section{Acknowledgements}
This work was supported by the National Key Research and Development Program of China (2022YFC3303600), National Natural Science Foundation of China (No. 62137002, 62293550, 62293553, 62293554, 62450005, 62437002, 62477036, 62477037, 62176209, 62192781, 62306229),  `LENOVO-XJTU' Intelligent Industry Joint Laboratory Project, the Shaanxi Provincial Social Science Foundation Project (No. 2024P041), the Natural Science Basic Research Program of Shaanxi (No. 2023-JC-YB-593), the National Research Foundation, Singapore, under its NRF Fellowship (Award\# NRF-NRFF14-2022-0001), the Youth Innovation Team of Shaanxi Universities `Multi-modal Data Mining and Fusion', Project of China Knowledge Centre for Engineering Science and Technology, and the Youth AI Talents Fund of China Association of Automation (Grant No. HBRC-JKYZD-2024-311). This work is supported by the Youth AI Talents Fund of China Association of Automation (Grant No.$\_$HBRC-JKYZD-2024-311. Mike Shou does not receive any funding for this work.

\appendix



\newpage
\section*{NeurIPS Paper Checklist}

\begin{enumerate}

\item {\bf Claims}
    \item[] Question: Do the main claims made in the abstract and introduction accurately reflect the paper's contributions and scope?
    \item[] Answer: \answerYes{} 
    \item[] Justification: We show the main claim in the abstract and introduction. And, we have concluded three contributions in Section \ref{sec:0}.

\item {\bf Limitations}
    \item[] Question: Does the paper discuss the limitations of the work performed by the authors?
    \item[] Answer: \answerYes{} 
    \item[] Justification: Section \ref{sec:5} explicitly points out that VLMs using CoFFT can correctly solve many problems that were previously unsolvable, but there are also some problems that can be solved correctly using VLM directly but CoFFT causes errors. 

\item {\bf Theory assumptions and proofs}
    \item[] Question: For each theoretical result, does the paper provide the full set of assumptions and a complete (and correct) proof?
    \item[] Answer: \answerNA{}.
    \item[] Justification: The paper does not include theoretical results.

    \item {\bf Experimental result reproducibility}
    \item[] Question: Does the paper fully disclose all the information needed to reproduce the main experimental results of the paper to the extent that it affects the main claims and/or conclusions of the paper (regardless of whether the code and data are provided or not)?
    \item[] Answer: \answerYes{}
    \item[] Justification: We have introduced the detailed settings such as implementation details and baselines in Section \ref{sec:4.1}. 

\item {\bf Open access to data and code}
    \item[] Question: Does the paper provide open access to the data and code, with sufficient instructions to faithfully reproduce the main experimental results, as described in supplemental material?
    \item[] Answer: \answerYes{}
    \item[] Justification: Yes, all codes will be open-sourced after the review process.

\item {\bf Experimental setting/details}
    \item[] Question: Does the paper specify all the training and test details (e.g., data splits, hyperparameters, how they were chosen, type of optimizer, etc.) necessary to understand the results?
    \item[] Answer: \answerYes{} 
    \item[] Justification: We have described the experimental details in Section \ref{sec:4.1}.

\item {\bf Experiment statistical significance}
    \item[] Question: Does the paper report error bars suitably and correctly defined or other appropriate information about the statistical significance of the experiments?
    \item[] Answer: \answerYes{} 
    \item[] Justification:  All experimental results as shown in Section \ref{sec:4.2} are the averages obtained after running three trials.

\item {\bf Experiments compute resources}
    \item[] Question: For each experiment, does the paper provide sufficient information on the computer resources (type of compute workers, memory, time of execution) needed to reproduce the experiments?
    \item[] Answer: \answerYes{} 
    \item[] Justification: All information about computer resources has been described in Section \ref{sec:4.1}.
    
\item {\bf Code of ethics}
    \item[] Question: Does the research conducted in the paper conform, in every respect, with the NeurIPS Code of Ethics \url{https://neurips.cc/public/EthicsGuidelines}?
    \item[] Answer: \answerYes{}
    \item[] Justification: This research complies with NeurIPS Code of Ethics.

\item {\bf Broader impacts}
    \item[] Question: Does the paper discuss both potential positive societal impacts and negative societal impacts of the work performed?
    \item[] Answer: \answerYes{}.
    \item[] Justification: This paper primarily focuses on technical performance impacts, showing positive improvements of 3.1\%-5.8\% on visual reasoning tasks, which can improve the practicality and reliability of current VLMs in complex visual reasoning scenarios, as shown in Section \ref{sec:4.2}.
    
\item {\bf Safeguards}
    \item[] Question: Does the paper describe safeguards that have been put in place for responsible release of data or models that have a high risk for misuse (e.g., pretrained language models, image generators, or scraped datasets)?
    \item[] Answer: \answerNA{}.
    \item[] Justification: This paper does not release any risky data.

\item {\bf Licenses for existing assets}
    \item[] Question: Are the creators or original owners of assets (e.g., code, data, models), used in the paper, properly credited and are the license and terms of use explicitly mentioned and properly respected?
    \item[] Answer: \answerYes{}
    \item[] Justification:  All benchmark and VLMs are the license and terms of use explicitly mentioned and properly respected in Section \ref{sec:4.1}.

\item {\bf New assets}
    \item[] Question: Are new assets introduced in the paper well documented and is the documentation provided alongside the assets?
    \item[] Answer: \answerYes{}.
    \item[] Justification: Yes, all assets will be publicly available after the review process.

\item {\bf Crowdsourcing and research with human subjects}
    \item[] Question: For crowdsourcing experiments and research with human subjects, does the paper include the full text of instructions given to participants and screenshots, if applicable, as well as details about compensation (if any)? 
    \item[] Answer: \answerNA{}.
    \item[] Justification: This paper does not involve crowdsourcing nor research with human subjects.

\item {\bf Institutional review board (IRB) approvals or equivalent for research with human subjects}
    \item[] Question: Does the paper describe potential risks incurred by study participants, whether such risks were disclosed to the subjects, and whether Institutional Review Board (IRB) approvals (or an equivalent approval/review based on the requirements of your country or institution) were obtained?
    \item[] Answer: \answerNA{}.
    \item[] Justification: This paper does not involve crowdsourcing nor research with human subjects.

\item {\bf Declaration of LLM usage}
    \item[] Question: Does the paper describe the usage of LLMs if it is an important, original, or non-standard component of the core methods in this research? Note that if the LLM is used only for writing, editing, or formatting purposes and does not impact the core methodology, scientific rigorousness, or originality of the research, declaration is not required.
    \item[] Answer: \answerYes{} 
    \item[] Justification: We use VLM to perform inference tests on multiple public benchmarks to observe the performance improvement of our approach, as shown in Section \ref{sec:4.1}.

\end{enumerate}

\bibliography{neurips_2025}

\newpage
\appendix
\section*{Appendix}
\section{Algorithmic Details}
\begin{algorithm}
\caption{Chain of Foresight-Focus Thought (CoFFT)}
\label{alg:cofft}
\begin{algorithmic}[1]
\Require Original image $V$, question $Q$, VLM model $M$, number of samples $k$, foresight length $l$, balancing factor $\lambda$, exploration factor $\alpha$, temperature range $[T_{\min}, T_{\max}]$
\Ensure Final reasoning result $R_n$ with answer

\State $V_0 \gets V$ \Comment{Initialize visual focus with original image}
\State $R_0 \gets \emptyset$ \Comment{Initialize reasoning chain as empty}
\State $t \gets 0$ \Comment{Initialize iteration counter}

\While{not reached final answer}
\State $t \gets t + 1$
\Comment{Stage 1: Diverse Sample Generation}
\State $S_t \gets \emptyset$ \Comment{Initialize empty sample set}
\For{$i \gets 1$ to $k$}
    \State $T_i \gets \text{Random}(T_{\min}, T_{\max})$ \Comment{Diverse temperature values}
    \State $s_i \gets \text{Generate}(M, V_{t-1}, Q, R_{t-1}, T_i, l)$ \Comment{Generate sample with $l$ steps}
    \State $S_t \gets S_t \cup \{s_i\}$
\EndFor

\Comment{Stage 2: Dual Foresight Decoding}
\For{each $s \in S_t$}
    \State $A^{rel}(V, s) \gets \text{Softmax}(\frac{A(V, s)}{A(V, D) + \epsilon})$ \Comment{Compute relative attention}
    \State $E_{att}(s) \gets 0.5 \cdot \cos(A^{rel}(V, Q), A^{rel}(V, s)) + 0.5 \cdot \text{IoU}^{30\%}(A^{rel}(V, Q), A^{rel}(V, s))$
    
    \State $p_0 \gets \text{MeanLogProb}(R_{t-1})$ \Comment{Base probability}
    \State $E_{prob}(s) \gets \frac{1}{l}\sum_{j=1}^{l}(\text{MeanLogProb}(R_{t-1} + s[1:j]) - p_0)$
\EndFor

\State $E_{att\_norm} \gets \text{Softmax}([E_{att}(s)] \; \forall s \in S_t)$
\State $E_{prob\_norm} \gets \text{Softmax}([E_{prob}(s)] \; \forall s \in S_t)$
\State $E_t \gets \lambda \cdot E_{att\_norm} + (1-\lambda) \cdot E_{prob\_norm}$
\State $s^* \gets \arg\max_{s \in S_t} E_t(s)$ \Comment{Select optimal sample}
\State $R_t \gets R_{t-1} \cup \{\text{FirstStep}(s^*)\}$ \Comment{Update reasoning with first step of optimal sample}

\Comment{Stage 3: Visual Focus Adjustment}
\State $A^{rel}(V, R_t) \gets \text{Softmax}(\frac{A(V, R_t)}{A(V, D) + \epsilon})$
\State $C^{rel}(V, Q, R_t) \gets \max(A^{rel}(V, Q) - \alpha \cdot A^{rel}(V, R_t), 0)$
\State $A_{crop} \gets 0.5 \cdot C^{rel}(V, Q, R_t) + 0.5 \cdot A^{rel}(V, s^*)$

\State $\Omega \gets \{\text{regions spanning 40\%-90\% of original dimensions}\}$
\State $\mu_{V_0} \gets \frac{1}{|V_0|}\sum_{(x,y) \in V_0} A_{crop}(x,y)$ \Comment{Global mean score}
\State $\sigma_{V_0} \gets \text{StandardDeviation}(A_{crop})$
\State $\beta \gets \sigma_{V_0} \cdot (1 - \cos(C^{rel}(V, Q, R_t), A^{rel}(V, s^*)))$

\State $B^* \gets \arg\max_{B \in \Omega} \frac{1}{|B|}\sum_{(x,y) \in B} A_{crop}(x,y)$ \Comment{Find optimal region}
\State $\mu_{B^*} \gets \frac{1}{|B^*|}\sum_{(x,y) \in B^*} A_{crop}(x,y)$ \Comment{Maximum region score}

\If{$\mu_{B^*} > \mu_{V_0} + \beta$}
    \State $V_t \gets \text{Crop}(V, B^*)$ \Comment{Crop and scale region}
\Else
    \State $V_t \gets V_0$ \Comment{Revert to original image}
\EndIf

\If{$R_t$ contains "REASONING\_COMPLETE"}
    \State \textbf{break}
\EndIf
\EndWhile

\State \Return $R_t$ \Comment{Return final reasoning with answer}
\end{algorithmic}
\end{algorithm}

\begin{algorithm}
\caption{Relative Attention Mechanism}
\label{alg:rel_attention}
\begin{algorithmic}[1]
\Require Image $V$, text input $X$, VLM model $M$, descriptive prompt $D$ (e.g., "Describe the image in detail")
\Ensure Relative attention map $A^{rel}(V, X)$

\State $A(V, X) \gets \text{ExtractAttentionMap}(M, V, X)$ \Comment{Extract raw attention map}
\State $A(V, D) \gets \text{ExtractAttentionMap}(M, V, D)$ \Comment{Extract descriptive attention map}
\State $A^{rel}(V, X) \gets \text{Softmax}(\frac{A(V, X)}{A(V, D) + \epsilon})$ \Comment{Compute relative attention}

\State \Return $A^{rel}(V, X)$
\end{algorithmic}
\end{algorithm}
Algorithm \ref{alg:cofft} presents the complete Chain of Foresight-Focus Thought (CoFFT) approach, which systematically enhances visual reasoning through an iterative cognitive-inspired process. 
The algorithm leverages three interconnected stages that collaborate to produce accurate reasoning paths.
\par
In the Diverse Sample Generation stage (lines 4-11), the algorithm explores different reasoning trajectories by generating $k$ diverse reasoning samples with varying temperature parameters. 
Each sample contains up to $l$ future reasoning steps, providing a meaningful lookahead that helps evaluate potential reasoning directions before committing to the next step. This diversity is crucial for addressing complex visual reasoning tasks where multiple valid reasoning paths may exist.
\par
The Dual Foresight Decoding stage (lines 12-22) represents the core evaluation mechanism, where samples are assessed through two complementary scores: a visual focus score ($E_{att}$) that measures visual relevance, and a reasoning progression score ($E_{prob}$) that evaluates logical coherence.
These scores are normalized and combined using a balancing factor $\lambda$, allowing the algorithm to select reasoning steps that are both visually grounded and logically sound. The first step of the optimal sample is then integrated into the evolving reasoning process.
\par
The Visual Focus Adjustment stage (lines 23-36) dynamically modifies visual attention based on current reasoning progress and future needs. 
It evaluates the significance of image regions using a dual-criteria scoring mechanism that considers both question relevance and future reasoning relevance.
The algorithm then employs a sliding window approach with adaptive thresholding to select optimal regions for focus.
This mechanism enables effective transitions between global context and local details throughout the reasoning process.
\par
The iterative cycle continues until a conclusive answer is reached (signaled by ``REASONING\_COMPLETE''), creating a synergistic loop where reasoning directs visual focus and optimized visual focus subsequently improves reasoning quality.
This cognitively-inspired approach significantly enhances the model's ability to perform complex visual reasoning tasks that require attention to both global context and local details.
\par
Algorithm \ref{alg:rel_attention} outlines the Relative Attention Mechanism, which serves as a foundational component for the visual focus calculations in Algorithm \ref{alg:cofft}. 
This concise mechanism addresses the challenge of redundant information in images by normalizing text-image attention maps against a baseline descriptive attention distribution.
\par
By performing element-wise division between task-specific attention and generic descriptive attention, followed by softmax normalization, this algorithm effectively highlights regions specifically relevant to the task while suppressing generally salient but task-irrelevant areas. 
This normalized attention map ($A^{rel}$) is then utilized throughout Algorithm \ref{alg:cofft} to compute visual focus scores and guide the visual focus adjustment process, creating a coherent system that intelligently navigates visual information during complex reasoning tasks.

\section{Implementation Details}
\subsection{Parameter Configuration}
The CoFFT algorithm requires several key parameters that influence its performance:
\begin{itemize}
\item \textbf{Number of samples ($k$)}: We set $k=4$ in our experiments, which balances computational efficiency with sufficient sample diversity.
\item \textbf{Foresight length ($l$)}: We use $l=5$ as the maximum number of future reasoning steps to consider, providing sufficient lookahead while maintaining computational efficiency.
\item \textbf{Balancing factor ($\lambda$)}: Set to 0.3, controlling the weight between visual focus score and reasoning progression score. This value was determined through ablation studies showing optimal performance.
\item \textbf{Exploration factor ($\alpha$)}: Set to 0.3, controlling how strongly the algorithm prioritizes unexplored regions over previously attended regions.
\item \textbf{Temperature range}: We use $T_{min}=0.4$ and $T_{max}=1.0$ with an interval of 0.1, resulting in a set of temperature values $\{0.4, 0.5, 0.6, 0.7, 0.8, 0.9, 1.0\}$ to ensure diversity in the generated reasoning samples.
\end{itemize}

\subsection{Temperature Sampling Strategy}
To ensure a comprehensive exploration of the reasoning space, we implemented an adaptive temperature sampling strategy:
\begin{itemize}
\item A temperature value is randomly selected from the set ${0.4, 0.5, 0.6, 0.7, 0.8, 0.9, 1.0}$ each time a sample is generated.
\item To prevent repeated selections and promote diversity, the probability weight of each chosen parameter is reduced by half in subsequent sampling processes.
\item The weights are reset to their initial values once all parameters have been selected, ensuring a balanced exploration of different temperature values throughout the reasoning process.
\end{itemize}

\subsection{Relative Attention Extraction}
The relative attention mechanism requires extracting attention maps from the VLM. For this purpose, we extract attention weights from the last few layers of the visual encoder and language decoder cross-attention modules. The attention maps are averaged across attention heads and layers to obtain a single attention distribution per text-image pair. These raw attention maps are then normalized according to Algorithm \ref{alg:rel_attention}.
\subsection{Stopping Criteria}
The iterative reasoning process continues until one of the following conditions is met:
\begin{itemize}
\item The model outputs the token ``REASONING\_COMPLETE'' indicating a final answer has been reached
\item The reasoning chain reaches a predefined maximum length (typically containing 3-7 reasoning steps, so we set this as 10)
\item The reasoning converges (negligible changes in subsequent iterations)
\end{itemize}

\section{Experimental Setup}
\subsection{Benchmarks}
Our evaluation encompassed multiple complementary benchmarks to assess various aspects of visual reasoning capabilities:
\begin{itemize}
\item \textbf{Mathematical and Geometric Reasoning}: MathVista \cite{lumathvista} and MathVision \cite{wang2024measuring}
\item \textbf{Multidisciplinary Visual Reasoning}: M3CoT \cite{chen2024m3cot} and MMStar \cite{chenwe}
\item \textbf{Chart Understanding}: Charxiv \cite{wang2024charxiv}
\item \textbf{Geographical Reasoning}: SeekWorld-Global (Google Maps panoramic imagery) and SeekWorld-China (Xiaohongshu App data) from the SeekWorld dataset \cite{seekworld2025}
\end{itemize}
\subsection{Compared Methods}
Our comparative analysis included the following state-of-the-art methods:
\begin{itemize}
\item \textbf{Search-based Method}: Monte Carlo Tree Search (MCTS) \cite{coulom2006efficient}
\item \textbf{Foresight Reasoning Method}: Predictive Decoding \cite{ma2024non}
\item \textbf{Visual Search Methodology}: DyFo \cite{li2025dyfo}
\item \textbf{Multi-modal Chain-of-thought Prompting}: ICoT \cite{gao2024interleaved}
\end{itemize}

\subsection{VLM Models}
Our experiments incorporated several state-of-the-art Vision Language Models:
\begin{itemize}
\item Qwen2.5-VL-Instruct (7B, 32B) \cite{bai2025qwen2}
\item InternVL2.5-Instruct (8B) \cite{wang2025internvideo2}
\item Llava-Next (7B) \cite{li2024llava}
\end{itemize}
These models were selected based on their architectural capabilities and demonstrated superior performance in visual reasoning tasks.
\subsection{Computational Infrastructure}
All experiments were executed on four NVIDIA A100 GPUs with parallel processing capabilities. For efficiency, sample generation and evaluation were parallelized across the GPUs to accelerate the experimentation process.
\subsection{Performance Metrics}
We adopted the following metrics for comprehensive evaluation:
\begin{itemize}
\item \textbf{Accuracy}: Pass@1 accuracy (Acc.) as the primary performance metric across all benchmarks
\item \textbf{Computational Efficiency}: Floating point operations (FLOPS) calculated following the methodology in \cite{kaplan2020scaling}, where $\text{FLOPS} \approx 6nP$ ($P$ represents model parameters, $n$ denotes generated tokens)
\end{itemize}
By computing the average number of tokens generated per example, we provided a standardized measure of computational cost across different methods based on Qwen2.5-VL-7B-Instruct to enable direct efficiency comparisons.

\section{Practical Implementation Notes}

\subsection{Region Selection Optimization}
The sliding window implementation was optimized using stride-based approaches rather than exhaustively evaluating all possible regions. We implemented an efficient algorithm that:

\begin{itemize}
\item Uses a stride of 10\% when scanning potential regions
\item Prioritizes regions with high attention density
\item Maintains aspect ratios within the original dimensions for subsequent reasoning iterations
\end{itemize}

\subsection{Inference Optimizations}
To maximize computational efficiency, we implemented several optimizations:

\begin{itemize}
\item \textbf{Batch Processing}: Sample generation and evaluation were parallelized across multiple GPUs
\item \textbf{Attention Caching}: Descriptive attention maps were cached to avoid redundant computation
\item \textbf{Early Stopping}: Inference was halted when ``REASONING\_COMPLETE'' was detected
\end{itemize}

This implementation provided a complete blueprint for deploying CoFFT in practice, demonstrating how our approach systematically addresses the challenges of complex visual reasoning through a cognitively-inspired iterative process.

\end{document}